\documentclass[12pt,a4paper]{article}

\usepackage[margin=1in]{geometry}
\usepackage{amsmath, amssymb}
\usepackage{graphicx}
\usepackage{booktabs}
\usepackage{array}
\usepackage{url}
\usepackage{hyperref}
\usepackage{authblk}
\usepackage{caption}
\usepackage{subcaption}
\usepackage{microtype}
\usepackage{setspace}
\usepackage{siunitx}
\usepackage{hyperref}
\usepackage{orcidlink} 

\hypersetup{
    colorlinks=true,
    linkcolor=blue,
    citecolor=blue,
    urlcolor=blue
}

\title{Rare Genomic Subtype Discovery from RNA-seq via Autoencoder Embeddings and Stability-Aware Clustering}

\author[1,*]{Alaa Mezghiche} 
\affil[1]{Department of Computer Science, University of Science and Technology Houari Boumediene (USTHB), Algiers, Algeria}
\affil[*]{Correspondence: \texttt{alaa.mezghiche@etu.usthb.dz}}
\date{}  

\begin{document}
\maketitle

\begin{abstract}
\noindent
Unsupervised learning on high-dimensional RNA-seq data can reveal molecular subtypes beyond standard labels. We combine an autoencoder-based representation with clustering and \emph{stability} analysis to search for \emph{rare} but reproducible genomic subtypes. On the UCI ``Gene Expression Cancer RNA-Seq'' dataset (801 samples, 20{,}531 genes; BRCA, COAD, KIRC, LUAD, PRAD), a \emph{pan-cancer} analysis shows clusters aligning almost perfectly with tissue-of-origin (Cram\'er’s $V = 0.887$), serving as a negative control. We therefore reframe the problem \emph{within} KIRC ($n = 146$): we select the top 2{,}000 highly variable genes, standardize them, train a feed-forward autoencoder (128-dimensional latent space), and run $k$-means for $k = 2 \dots 10$. While global indices favor small $k$, scanning $k$ with a pre-specified discovery rule (rare $< 10\%$ and stable with Jaccard $\ge 0.60$ across 20 seeds after Hungarian alignment) yields a simple solution at $k = 5$ (silhouette $= 0.129$, DBI $= 2.045$) with a rare cluster C0 (6.85\% of patients) that is highly stable (Jaccard $= 0.787$). Cluster-vs-rest differential expression (Welch’s $t$-test, Benjamini--Hochberg FDR) identifies coherent markers, including strong down-regulation of \texttt{gene\_11713}, \texttt{gene\_13678}, \texttt{gene\_16402}, \texttt{gene\_3777} and up-regulation of \texttt{gene\_751}, \texttt{gene\_17397}, \texttt{gene\_2760}. Overall, pan-cancer clustering is dominated by tissue-of-origin, whereas a stability-aware within-cancer approach reveals a rare, reproducible KIRC subtype.
\end{abstract}


\section{Introduction}

High-throughput RNA sequencing (RNA-seq) has transformed our ability to profile gene expression at scale, enabling large consortia such as The Cancer Genome Atlas (TCGA) to generate transcriptomic data across many tumor types and patients. Unsupervised analysis of such data has the potential to uncover \emph{molecular subtypes}, or groups of patients sharing coherent expression programs that may not correspond directly to current histopathological classifications.

The UCI ``Gene Expression Cancer RNA-Seq'' dataset is a curated subset (801 patients, 20{,}531 genes) of the TCGA Pan-Cancer RNA-seq collection, covering five tumor types: breast invasive carcinoma (BRCA), colon adenocarcinoma (COAD), kidney renal clear cell carcinoma (KIRC), lung adenocarcinoma (LUAD) and prostate adenocarcinoma (PRAD)~\cite{fiorini2016}. This dataset has been widely used as a benchmark for supervised cancer classification and feature selection, including autoencoder-based biomarker identification~\cite{alabir2022} and comparative studies of machine learning models on RNA-seq data.

Most existing work on this dataset focuses on predicting the known tumor type from gene expression. In contrast, our goal is to explore whether unsupervised methods can discover \emph{rare} transcriptomic subtypes within a given cancer type. Such rare subtypes, if stable and biologically coherent, could correspond to clinically meaningful subgroups (e.g., distinct prognostic profiles or therapy responses).

Autoencoders provide a natural way to compress high-dimensional gene expression into a lower-dimensional latent space while preserving non-linear structure. Clustering in this latent space, combined with interpretability tools, can help reveal underlying patterns~\cite{alabir2022}. However, pan-cancer clustering is strongly driven by tissue-of-origin, which can mask finer-grained within-cancer structure. 

In this work, we:

\begin{enumerate}
    \item Perform a pan-cancer unsupervised analysis as a sanity / negative control, confirming that clusters align with tissue labels and therefore mostly re-discover known structure.
    \item Reframe the task within a single cancer type (KIRC), using an autoencoder to learn a compact representation of highly variable genes.
    \item Use $k$-means clustering with model selection and cluster stability analysis to identify rare but reproducible clusters.
    \item Apply simple differential expression analysis (cluster-vs-rest) to derive a cluster-specific gene signature.
\end{enumerate}

\section{Materials and Methods}

\subsection{Dataset}

We use the ``Gene Expression Cancer RNA-Seq'' dataset from the UCI Machine Learning Repository~\cite{fiorini2016}. This dataset consists of:
\begin{itemize}
    \item $801$ tumor samples (patients).
    \item $20{,}531$ gene expression features measured by Illumina HiSeq RNA-seq.
    \item Five tumor types (``Class'' label): BRCA ($n=300$), KIRC ($n=146$), LUAD ($n=141$), PRAD ($n=136$), COAD ($n=78$).
\end{itemize}

Expression values are provided in a matrix where rows are samples and columns are genes (named \texttt{gene\_0} to \texttt{gene\_20530}). Class labels are provided separately and aligned to samples by an ID field (\texttt{sample\_X}). We merged features and labels using an inner join on the sample ID, and subsequently set the sample ID as the row index.

\subsection{Pan-cancer negative control}

As an initial experiment, we considered all 801 samples together (pan-cancer analysis). The objective was to confirm that the unsupervised pipeline is sensible and that the latent space captures known structure.

\paragraph{Preprocessing.}
We extracted the gene expression matrix $X \in \mathbb{R}^{801 \times 20531}$ and verified that all values were finite and non-negative. We applied a $\log(1 + x)$ transform to reduce skewness. To reduce noise and dimensionality, we selected the top 2{,}000 most variable genes across all samples based on the empirical variance, then standardized each gene to zero mean and unit variance (z-scoring).

\begin{figure}[h!]
    \centering
    \includegraphics[width=0.7\textwidth]{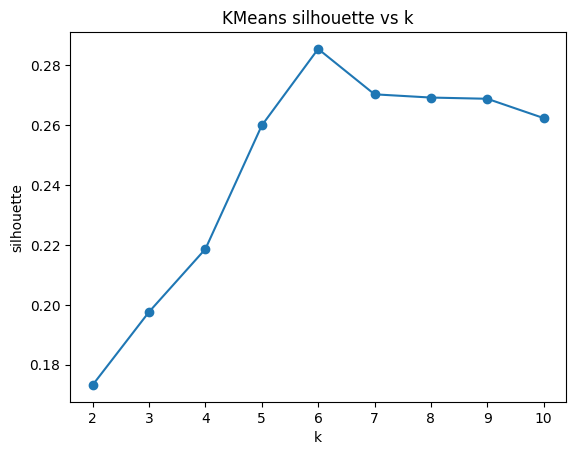}
    \caption{Pan-cancer k-means silhouette score as a function of $k$, computed on the autoencoder latent space ($Z$) for all 801 samples. The maximum occurs at $k=6$.}
    \label{fig:pan_silhouette}
\end{figure}

\paragraph{Pan-cancer autoencoder and clustering.}
We trained a feed-forward autoencoder on the standardized matrix $X_{\text{scaled}} \in \mathbb{R}^{801 \times 2000}$ and obtained latent codes $Z \in \mathbb{R}^{801 \times 128}$ (architecture described in Section~\ref{sec:ae_arch}). We then applied $k$-means clustering on $Z$ for $k=2,\dots,10$ and computed the silhouette score for each $k$, which increases from $0.174$ at $k=2$ to a maximum of $0.286$ at $k=6$ before slightly decreasing (Fig.~\ref{fig:pan_silhouette}). We chose $k=6$ for detailed inspection as a reasonable pan-cancer solution.

\paragraph{Evaluation.}
We constructed a contingency table between tumor types and clusters, row-normalized it, and measured association using Pearson’s chi-square test and Cram\'er’s $V$.

\subsection{Within-cancer reframing: focus on KIRC}

To avoid the dominant tissue-of-origin signal, we restricted the analysis to one cancer type: KIRC (kidney renal clear cell carcinoma). We filtered the merged dataset to keep only samples with \texttt{Class=KIRC}, yielding $n = 146$ patients. All subsequent analyses were performed on this subset.

\subsubsection{Highly variable genes and scaling}

Within KIRC, we computed the variance of each gene across the 146 samples and selected the top 2{,}000 most variable genes (or fewer if fewer genes had non-zero variance). Let $X_{\text{KIRC}} \in \mathbb{R}^{146 \times G}$ denote the KIRC expression matrix for these genes, with $G \leq 2000$.

We standardized each gene to zero mean and unit variance:
\[
X_{z} = \text{StandardScaler}(X_{\text{KIRC}}),
\]
resulting in $X_{z} \in \mathbb{R}^{146 \times G}$, which serves as input to the autoencoder.

\subsection{Autoencoder architecture and training}
\label{sec:ae_arch}

We used the same architectural template for both pan-cancer and KIRC analyses, instantiated with the appropriate input dimension ($D_{\text{in}} = 2000$ for pan-cancer; $D_{\text{in}} = G$ within KIRC).

The autoencoder is a fully connected neural network implemented in PyTorch:

\begin{itemize}
    \item Input dimension: $D_{\text{in}}$ (number of selected genes).
    \item Hidden layer 1: size $H_1 = \min(1024, \max(256, D_{\text{in}}/2))$, with ReLU activation and dropout ($p=0.1$).
    \item Hidden layer 2: size $H_2 = \min(512, \max(128, D_{\text{in}}/4))$, with ReLU activation.
    \item Latent layer: size $D_{\text{lat}} = 128$ (linear).
    \item Decoder: symmetric to the encoder with ReLU activations.
\end{itemize}

The encoder $f_\theta: \mathbb{R}^{D_{\text{in}}} \to \mathbb{R}^{128}$ maps input samples to latent codes $z$, and the decoder reconstructs back to input space. We trained the model using mean squared error (MSE) reconstruction loss:
\[
\mathcal{L} = \frac{1}{N} \sum_{i=1}^{N} \| x_i - \hat{x}_i \|_2^2,
\]
with the Adam optimizer (learning rate $10^{-3}$, weight decay $10^{-5}$). We used mini-batches of size 256, a 85\%/15\% train/validation split, and early stopping (patience $=15$).

\begin{figure}[h!]
    \centering
    \includegraphics[width=0.7\textwidth]{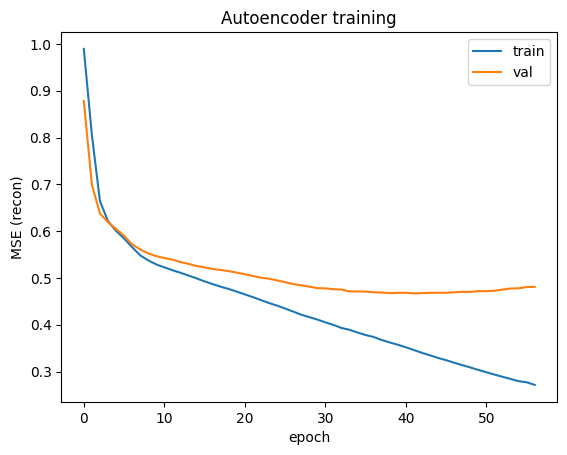}
    \caption{Example autoencoder training curve (KIRC): reconstruction MSE vs epoch for training and validation splits.}
    \label{fig:ae_training}
\end{figure}

A typical learning curve for the KIRC autoencoder is shown in Fig.~\ref{fig:ae_training}: training and validation MSE decrease rapidly in the first 5 epochs, then flatten; the validation curve stabilizes around $0.46$ while the training curve continues to decrease more slowly.

After training, we embedded \emph{all} KIRC samples by passing $X_{z}$ through the encoder in evaluation mode (without dropout), obtaining a latent matrix $Z \in \mathbb{R}^{146 \times 128}$.

\subsection{Clustering, model selection and stability}

\subsubsection{Within-KIRC clustering}

Within KIRC, we applied $k$-means clustering on the latent codes $Z$ for $k$ in $\{2,\dots,10\}$, using $n_{\text{init}} = 10$ and a fixed random seed (42). For each $k$, we computed:

\begin{itemize}
    \item The silhouette score (higher is better).
    \item The Davies--Bouldin index (DBI; lower is better).
\end{itemize}

\begin{figure}[h!]
    \centering
    \includegraphics[width=0.7\textwidth]{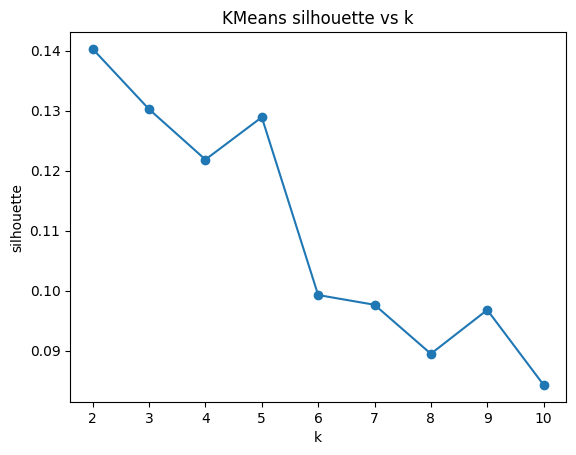}
    \caption{Within-KIRC k-means silhouette score as a function of $k$. The best score is at $k=2$, but rare subtypes emerge only at larger $k$.}
    \label{fig:kirc_silhouette}
\end{figure}

\begin{figure}[h!]
    \centering
    \includegraphics[width=0.7\textwidth]{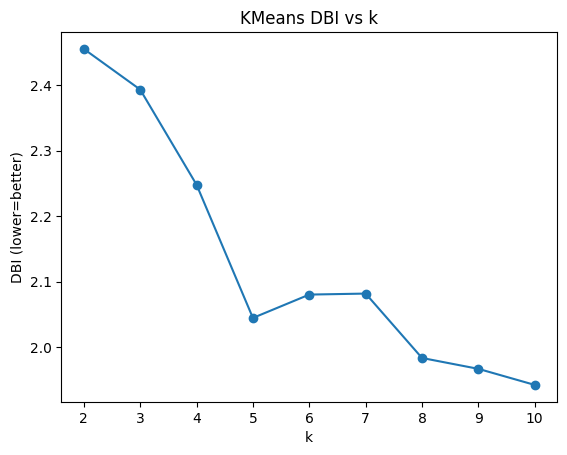}
    \caption{Within-KIRC Davies--Bouldin index (DBI) as a function of $k$. Lower values indicate more compact and separated clusters.}
    \label{fig:kirc_dbi}
\end{figure}

The silhouette curve peaks at $k=2$ with a score of $0.140$, then slowly declines as $k$ increases (Fig.~\ref{fig:kirc_silhouette}). DBI also decreases gradually with $k$ and reaches $2.045$ at $k=5$ before continuing to drop (Fig.~\ref{fig:kirc_dbi}). The purely ``best'' solution in terms of these global indices is therefore a small $k$ (particularly $k=2$), but this produces only two large clusters (sizes 69 and 77) with no rare subtypes.

To explicitly search for rare but meaningful clusters, we supplemented these global metrics with a stability analysis described below.

\subsubsection{Cluster stability via Jaccard index}

For each value of $k$, we ran $k$-means $R=20$ times with different random seeds, obtaining labelings $\ell^{(1)}, \dots, \ell^{(R)}$. Because cluster labels are arbitrary up to permutation, we aligned all runs to a reference labeling (e.g., $\ell^{(1)}$) using the Hungarian algorithm.

For each cluster $c$ in the reference solution, we computed its Jaccard similarity across runs:
\[
J_c = \frac{1}{R-1} \sum_{r=2}^{R} \frac{|S_c \cap S_c^{(r)}|}{|S_c \cup S_c^{(r)}|},
\]
where $S_c$ is the set of samples assigned to cluster $c$ in the reference run, and $S_c^{(r)}$ is the aligned set in run $r$.

We then summarized, for all $k$ and clusters, the triplet:
\begin{itemize}
    \item Prevalence $p_c = |S_c| / N$,
    \item Jaccard stability $J_c$,
    \item Indicators for being \emph{rare} ($p_c < 0.10$) and \emph{stable} ($J_c \ge 0.60$).
\end{itemize}

Across all $k$, rare \emph{and} stable clusters appear at:
\[
(k, \text{cluster}) \in \{ (5,0), (8,5), (10,7), (10,9) \}.
\]
Among these, the configuration with $k=5$ is the simplest and has the largest global silhouette (0.129) and reasonably low DBI (2.045). We therefore focus on the KIRC clustering solution with $k=5$.

\subsubsection{Final clustering solution at $k=5$}

For $k=5$, we refitted $k$-means with $n_{\text{init}}=30$ to obtain stable labels $c_1,\dots,c_{146} \in \{0,1,2,3,4\}$. The resulting cluster sizes and prevalences are:
\[
[10, 19, 31, 53, 33]
\quad\leftrightarrow\quad
[6.85\%, 13.0\%, 21.2\%, 36.3\%, 22.6\%].
\]

Cluster C0 is the rare cluster with size 10 (6.85\%). The global clustering metrics at $k=5$ are:
\[
\text{silhouette} = 0.129, \qquad \text{DBI} = 2.045.
\]

\begin{figure}[h!]
    \centering
    \includegraphics[width=0.7\textwidth]{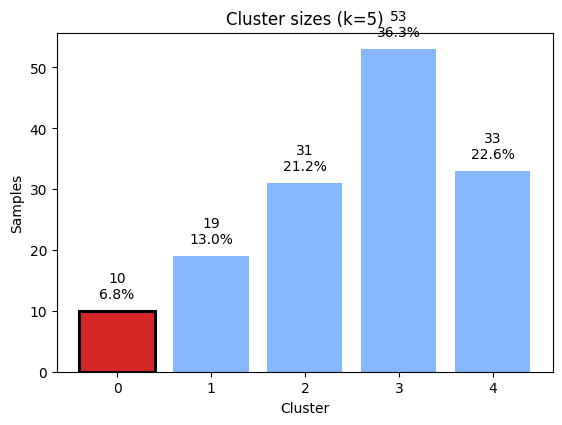}
    \caption{Cluster sizes for KIRC at $k=5$. Cluster C0 (size 10, prevalence 6.8\%) is highlighted in red as the rare subtype.}
    \label{fig:cluster_sizes}
\end{figure}

\begin{figure}[h!]
    \centering
    \includegraphics[width=0.7\textwidth]{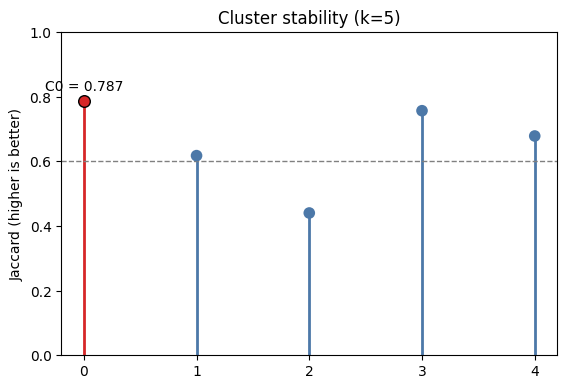}
    \caption{Cluster stability for KIRC at $k=5$, measured by Jaccard index across 20 random seeds after Hungarian alignment. C0 (red) has Jaccard 0.787 (above the 0.60 stability threshold, dashed line).}
    \label{fig:stability}
\end{figure}

The barplot in Fig.~\ref{fig:cluster_sizes} highlights C0 in red and shows the distribution of cluster sizes. Stability analysis for $k=5$ (Fig.~\ref{fig:stability}) confirms that C0 is highly stable across runs with a Jaccard index of $0.787$, while other clusters have Jaccard indices between $0.44$ and $0.76$.

\subsection{Cluster interpretability via differential expression}

To characterize the rare KIRC cluster C0, we performed a cluster-vs-rest differential expression (DE) analysis on the standardized KIRC expression matrix $X_{z}$. For C0 (the 10 in-cluster samples) and the remaining 136 out-of-cluster samples, we computed for each gene $g$:

\begin{itemize}
    \item The effect size: $\Delta z_g = \mu_{g,\text{in}} - \mu_{g,\text{out}}$.
    \item A Welch’s $t$-test $p$-value comparing in-cluster vs out-of-cluster samples.
    \item Benjamini--Hochberg FDR correction over all genes.
\end{itemize}

\begin{figure}[h!]
    \centering
    \includegraphics[width=0.7\textwidth]{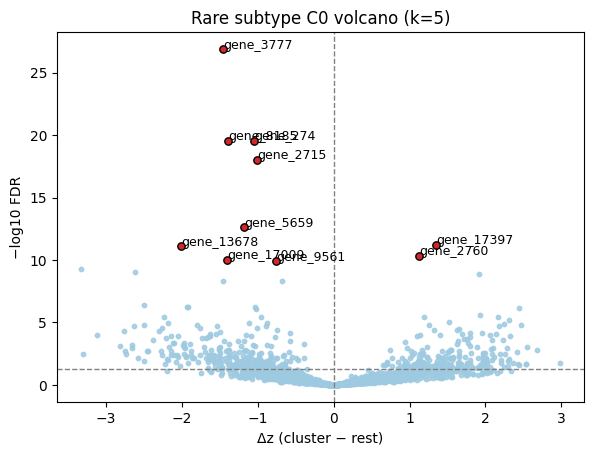}
    \caption{Volcano plot for C0 (cluster-vs-rest DE on standardized expression). Each point is a gene; red labelled points denote the top markers listed in Table~\ref{tab:markers}.}
    \label{fig:volcano}
\end{figure}

\begin{figure}[h!]
    \centering
    \includegraphics[width=\textwidth]{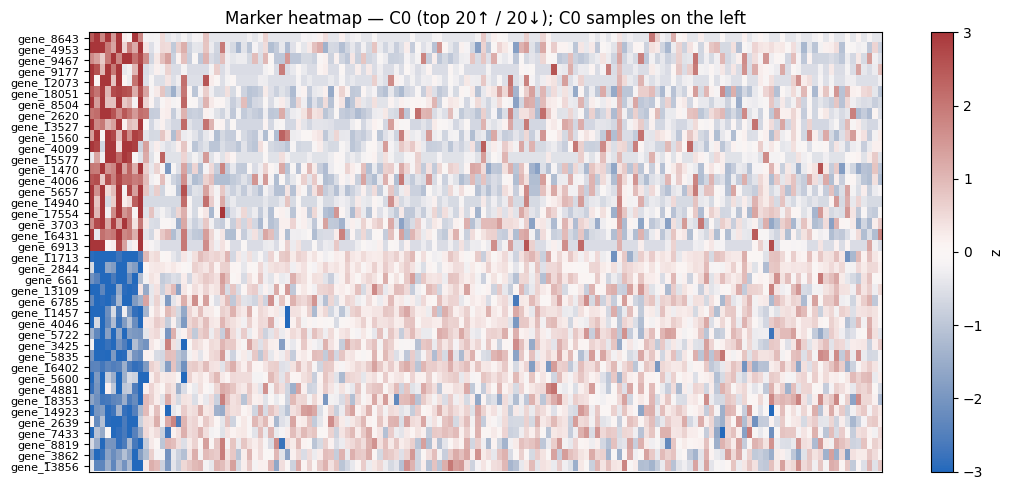}
    \caption{Heatmap of standardized expression ($z$-scores) for the top 20 up-regulated and top 20 down-regulated genes in C0. Samples are ordered with C0 on the left.}
    \label{fig:heatmap}
\end{figure}

The top hits (sorted by FDR and $|\Delta z_g|$) are reported in Table~\ref{tab:markers}. A volcano plot (Fig.~\ref{fig:volcano}) shows the global distribution of genes; those with large $|\Delta z_g|$ and low FDR are annotated. A heatmap of the top 20 up- and top 20 down-regulated genes (Fig.~\ref{fig:heatmap}) confirms that C0 patients share a coherent molecular pattern distinct from other KIRC samples.

\subsection{Visualization of latent structure}

\begin{figure}[h!]
    \centering
    \includegraphics[width=0.6\textwidth]{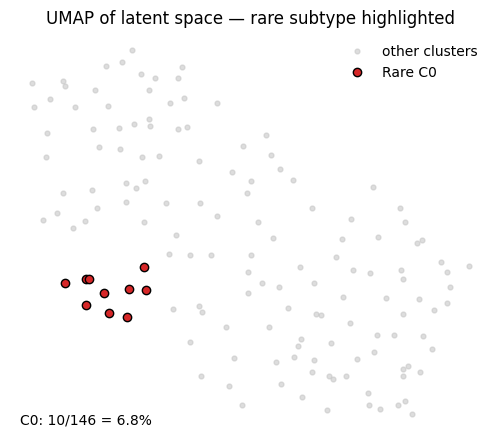}
    \caption{UMAP of the KIRC latent space, with rare cluster C0 highlighted in red and other clusters in grey. C0 occupies a compact region of the latent space.}
    \label{fig:umap}
\end{figure}

To visualize the structure of the latent space, we applied UMAP to the KIRC latent codes $Z$:
\[
U = \text{UMAP}(n\_neighbors = 15, \ \text{min\_dist} = 0.3, \ \text{random\_state} = 42).
\]
The resulting 2D embedding (Fig.~\ref{fig:umap}) shows that C0 samples are concentrated in a compact region of the latent space, separated from the bulk of other KIRC samples, further supporting the idea that C0 represents a coherent subtype rather than scattered noise.

\section{Results}

\subsection{Pan-cancer clusters re-discover tissue-of-origin}

For the pan-cancer analysis, the $k=6$ solution yields the cluster size and stability profile shown in Table~\ref{tab:pan_summary}. All clusters are extremely stable (Jaccard indices $\ge 0.994$), including two rare clusters (prevalence $<10\%$).

\begin{table}[h!]
\centering
\caption{Pan-cancer k-means clusters at $k=6$.}
\label{tab:pan_summary}
\begin{tabular}{lrrrrr}
\toprule
Cluster & Size & Prevalence & Jaccard & Rare ($<10\%$) & Stable ($\ge 0.60$) \\
\midrule
0 & 240 & 0.300 & 0.998 & No & Yes \\
1 & 136 & 0.170 & 1.000 & No & Yes \\
2 &  65 & 0.081 & 0.994 & Yes & Yes \\
3 & 136 & 0.170 & 1.000 & No & Yes \\
4 & 145 & 0.181 & 1.000 & No & Yes \\
5 &  79 & 0.099 & 1.000 & Yes & Yes \\
\bottomrule
\end{tabular}
\end{table}

The contingency table between tumor type (rows) and cluster (columns) is:

\begin{center}
\begin{tabular}{lrrrrrr}
\toprule
Class & 0 & 1 & 2 & 3 & 4 & 5 \\
\midrule
BRCA & 240 &   0 & 60 &   0 &   0 &   0 \\
COAD &   0 &   0 &  0 &   0 &   0 &  78 \\
KIRC &   0 &   0 &  1 &   0 & 145 &   0 \\
LUAD &   0 &   0 &  4 & 136 &   0 &   1 \\
PRAD &   0 & 136 &  0 &   0 &   0 &   0 \\
\bottomrule
\end{tabular}
\end{center}

Row-normalizing this table gives:

\begin{center}
\begin{tabular}{lrrrrrr}
\toprule
Class & 0 & 1 & 2 & 3 & 4 & 5 \\
\midrule
BRCA & 0.80 & 0.00 & 0.20 & 0.00 & 0.00 & 0.00 \\
COAD & 0.00 & 0.00 & 0.00 & 0.00 & 0.00 & 1.00 \\
KIRC & 0.00 & 0.00 & 0.01 & 0.00 & 0.99 & 0.00 \\
LUAD & 0.00 & 0.00 & 0.03 & 0.96 & 0.00 & 0.01 \\
PRAD & 0.00 & 1.00 & 0.00 & 0.00 & 0.00 & 0.00 \\
\bottomrule
\end{tabular}
\end{center}

The chi-square test of independence between tumor type and cluster assignment is highly significant ($p \approx 0$), with Cram\'er’s $V = 0.887$. Thus, the pan-cancer clusters essentially recover the five tissues-of-origin:

\begin{itemize}
    \item Cluster 0: mostly BRCA (240/300).
    \item Cluster 1: pure PRAD (136/136).
    \item Cluster 3: almost pure LUAD (136/141, plus a few LUAD samples in other clusters).
    \item Cluster 4: almost pure KIRC (145/146).
    \item Cluster 5: almost pure COAD (78/78, plus 1 LUAD).
    \item Cluster 2: a small residual cluster (65 samples) with the remaining BRCA and a few LUAD/KIRC samples.
\end{itemize}

Although there are two rare clusters (2 and 5), they correspond to leftover fractions of known tumour types rather than novel cross-cancer subtypes. We therefore treat this experiment as a \emph{negative control} that validates the pipeline but does not produce surprising biology.

\subsection{Within KIRC: rare and stable subtype C0 at $k=5$}

In contrast, the within-KIRC analysis allows us to search for more subtle heterogeneity. As noted above, pure clustering metrics favour $k=2$ (silhouette $0.140$, DBI $2.455$), but this solution contains only two large clusters:

\begin{center}
\begin{tabular}{lrrr}
\toprule
Cluster & Size & Prevalence & Rare \\
\midrule
0 & 69 & 0.473 & No \\
1 & 77 & 0.527 & No \\
\bottomrule
\end{tabular}
\end{center}

By scanning $k$ and examining cluster stability, we identify several rare and stable clusters (Table in the Methods section). Among them, cluster C0 at $k=5$ has:

\begin{itemize}
    \item Size $=10$ patients ($6.85\%$ of KIRC).
    \item Jaccard stability $=0.787$ across 20 seeds.
    \item Clear separation in the UMAP latent space (Fig.~\ref{fig:umap}).
\end{itemize}

Therefore, we focus on this $k=5$ solution and interpret C0 as a candidate rare genomic subtype of KIRC.

\subsection{Differential expression and marker genes of C0}

The DE analysis for C0 identifies a set of strongly altered genes. The top 15 markers are summarized in Table~\ref{tab:markers} (values taken directly from the analysis):

\begin{table}[h!]
\centering
\caption{Top 15 marker genes for the rare KIRC cluster C0 (cluster-vs-rest DE on standardized expression). Negative effects indicate down-regulation in C0; positive effects indicate up-regulation.}
\label{tab:markers}
\begin{tabular}{lrrr}
\toprule
Gene & Effect $\Delta z$ & $p$-value & FDR \\
\midrule
\texttt{gene\_3777}  & $-1.459$ & $6.34 \times 10^{-31}$ & $1.27 \times 10^{-27}$ \\
\texttt{gene\_274}   & $-1.050$ & $3.38 \times 10^{-23}$ & $2.73 \times 10^{-20}$ \\
\texttt{gene\_8185}  & $-1.386$ & $4.10 \times 10^{-23}$ & $2.73 \times 10^{-20}$ \\
\texttt{gene\_2715}  & $-1.013$ & $1.80 \times 10^{-21}$ & $9.01 \times 10^{-19}$ \\
\texttt{gene\_5659}  & $-1.178$ & $5.86 \times 10^{-16}$ & $2.35 \times 10^{-13}$ \\
\texttt{gene\_17397} & $+1.351$ & $1.75 \times 10^{-14}$ & $5.84 \times 10^{-12}$ \\
\texttt{gene\_13678} & $-2.012$ & $2.84 \times 10^{-14}$ & $8.10 \times 10^{-12}$ \\
\texttt{gene\_2760}  & $+1.131$ & $1.82 \times 10^{-13}$ & $4.54 \times 10^{-11}$ \\
\texttt{gene\_17009} & $-1.406$ & $4.11 \times 10^{-13}$ & $9.12 \times 10^{-11}$ \\
\texttt{gene\_9561}  & $-0.755$ & $6.57 \times 10^{-13}$ & $1.31 \times 10^{-10}$ \\
\texttt{gene\_11713} & $-3.332$ & $2.85 \times 10^{-12}$ & $5.19 \times 10^{-10}$ \\
\texttt{gene\_16402} & $-2.612$ & $5.92 \times 10^{-12}$ & $9.87 \times 10^{-10}$ \\
\texttt{gene\_751}   & $+1.915$ & $9.12 \times 10^{-12}$ & $1.40 \times 10^{-9}$ \\
\texttt{gene\_17921} & $-0.674$ & $3.17 \times 10^{-11}$ & $4.53 \times 10^{-9}$ \\
\texttt{gene\_5945}  & $-1.457$ & $3.80 \times 10^{-11}$ & $5.07 \times 10^{-9}$ \\
\bottomrule
\end{tabular}
\end{table}

The volcano plot (Fig.~\ref{fig:volcano}) highlights these genes as red points with labels; they occupy the extremes of the $\Delta z$ axis and the top of the $-\log_{10}(\text{FDR})$ axis, confirming both strong effect size and statistical significance. The heatmap (Fig.~\ref{fig:heatmap}) of the top 20 up- and 20 down-regulated genes (with C0 samples placed on the left) shows a clear block structure: C0 samples share strong down-regulation of several genes (deep blue) and up-regulation of a smaller set (red), while the rest of the KIRC cohort shows more heterogeneous expression.

\section{Discussion}

This work presents a small but illustrative pipeline for discovering rare transcriptomic subtypes in cancer using autoencoders, clustering and simple statistical interpretability. Using the UCI Gene Expression Cancer RNA-Seq dataset as a testbed, we show that:

\begin{enumerate}
    \item Pan-cancer unsupervised clustering primarily re-discovers tissue-of-origin, with clusters essentially corresponding to BRCA, COAD, KIRC, LUAD and PRAD. The high Cram\'er’s $V$ value and nearly block-diagonal contingency table confirm that the model captures known structure but does not reveal unexpected cross-cancer subtypes in this dataset.
    \item Restricting to a single cancer type (KIRC) and focusing on highly variable genes allows us to search for finer-grained within-cancer heterogeneity without the confounding effect of tissue differences.
    \item An autoencoder-based latent representation, combined with $k$-means clustering and Jaccard-based stability analysis, can highlight rare but robust clusters. In our case, cluster C0 represents only $\sim 7\%$ of KIRC patients but remains stable across multiple random seeds.
    \item Differential expression analysis yields a compact list of marker genes for the rare cluster, suggesting that it corresponds to a distinct transcriptional state rather than noise.
\end{enumerate}

Because the UCI dataset uses anonymized gene identifiers (\texttt{gene\_X}) and aggregates samples from a larger TCGA resource, we cannot directly map our markers to known gene symbols or pathways in this setting. As a result, our current analysis is methodological: it demonstrates that the combination of autoencoders, clustering, and stability analysis can identify candidate rare subtypes, but does not yet provide biological interpretation.

A natural next step is to re-run the same pipeline on the full TCGA KIRC RNA-seq data where genes have standard identifiers. This would enable downstream analyses such as Gene Ontology and pathway enrichment, comparison with known KIRC subtypes, and correlation with clinical variables such as survival, stage, and treatment.

\section{Conclusion}

We presented an unsupervised pipeline for discovering rare genomic subtypes within cancer using autoencoders, clustering, cluster stability analysis, and differential expression. Applied to the KIRC subset of the UCI Gene Expression Cancer RNA-Seq dataset, this approach identified a rare ($\sim 7\%$) but stable cluster with a distinct expression signature.

While limited by anonymized gene identifiers and lack of clinical annotations in the benchmark dataset, this work provides a proof-of-concept that can be extended to richer datasets such as TCGA with full clinical and genomic annotations. Ultimately, combining unsupervised representation learning, stability-aware clustering and interpretable signatures may help uncover clinically actionable subgroups in cancer.

\section*{Data and code availability}

All expression data used in this study are available from the UCI Machine Learning Repository under the title ``Gene Expression Cancer RNA-Seq'' (dataset ID 401)~\cite{fiorini2016}. 
The analysis code (autoencoder training, clustering, stability analysis and differential expression) is implemented in Python (PyTorch, scikit-learn, statsmodels) and is available at:
\url{https://github.com/alaa-32/Discovering-Rare-Genomic-Subtypes-from_RNA-seq.git}.

\bibliographystyle{unsrt}


\clearpage

\end{document}